# Diacritic Recognition Performance in Arabic ASR

*Hanan Aldarmaki, Ahmad Ghannam*

MBZUAI

`hanan.aldarmaki@mbzuai.ac.ae`, `ahmad.ghannam@mbzuai.ac.ae`

## Abstract

We present an analysis of diacritic recognition performance in Arabic Automatic Speech Recognition (ASR) systems. As most existing Arabic speech corpora do not contain all diacritical marks, which represent short vowels and other phonetic information in Arabic script, current state-of-the-art ASR models do not produce full diacritization in their output. Automatic text-based diacritization has previously been employed both as a pre-processing step to train diacritized ASR, or as a post-processing step to diacritize the resulting ASR hypotheses. It is generally believed that input diacritization degrades ASR performance, but no systematic evaluation of ASR diacritization performance, independent of ASR performance, has been conducted to date. In this paper, we attempt to experimentally clarify whether input diacritiztation indeed degrades ASR quality, and to compare the diacritic recognition performance against text-based diacritization as a post-processing step. We start with pre-trained Arabic ASR models and fine-tune them on transcribed speech data with different diacritization conditions: manual, automatic, and no diacritization. We isolate diacritic recognition performance from the overall ASR performance using coverage and precision metrics. We find that ASR diacritization significantly outperforms text-based diacritization in post-processing, particularly when the ASR model is fine-tuned with manually diacritized transcripts.

**Index Terms**: arabic speech recognition, automatic diacritization

## 1. Introduction

Arabic diacritics are small marks placed above or below alphabetical characters to indicate additional information, such as short vowels that are not represented in the Arabic alphabet, as well as gemination (i.e. consonant doubling) and some pronounceable syntactic marks. However, due to their peripheral presence, most people write and type Arabic text without the inclusion of diacritics. At best, partial diacritics are sometimes added in particularly ambiguous cases, but most diacritics are omitted from text and left to be inferred from context. Special texts, like religious scripture or introductory Arabic learning material, may contain full diacritics. Some other resources are manually diacritized for research and development purposes. For example, the Tashkeela corpus [1] contains 55K[1] manually diacritized sentences and is commonly used to train automatic diacritization models. Similarly, most speech corpora do not include diacritics in their transcriptions, except if they are recitations of religious text (e.g. the Quranic Arabic Corpus[2]) or if they are curated for text-to-speech applications (e.g. the Arabic Speech Corpus [3][3]). State-of-the-art ASR models are typically trained with combinations of different speech data sets with a mixture of diacritization conditions. As a result, ASR outputs tend to have low coverage of diacritics, and the coverage depends on context. For instance, we observed that the pre-trained Whisper ASR model[4] produces full diacritics for some but not all Quaranic verses, and almost no diacritics for casual MSA speech.

The omission of diacritics in text has two opposite effects in machine learning models: reducing sparsity and increasing lexical ambiguity. As many words with different pronunciation and meaning end up with the same text transcription, omitting diacritics leads to an increase in the number of homographs that can be difficult to disambiguate. On the other hand, keeping full diacritics often results in sparsity effects, where some word variants are observed less frequently or never, which leads to out-of-vocabulary and generalization errors. Intermediate levels of diacritizations can be used as a compromise (see for example, [4] and [5]), but partial diacritization hasn't been widely adopted in automated systems due to the subjective nature of annotations. In stand-alone ASR, lexical ambiguity in the output space is less of a concern compared to applications where text is used as input. Without consideration of further post-processing steps or the possible use cases of the ASR output, it can seem reasonable to omit diacritics in the transcriptions to simplify the output space and minimize the effects of sparsity.

Whether diacritics in ASR output are desirable or not depends on their intended use or the training conditions of downstream applications (e.g. consider an application where ASR output is used as input to a machine translation system). If the downstream model is trained with undiacritized input text, the output of ASR will be post-processed by removing all diacritics, if any. If the downstream application is trained with full diacritics, on the other hand, the output of ASR will need to be fully diacritized. An ad-hoc solution in the latter case is to restore the diacritics as a post-processing step using a text-based diacritizer. However, we contend that diacritics produced directly from the ASR system have the potential to be more accurate than text-based diacritizers: while text-based models rely exclusively on textual context, a speech model has access to the original audio signal which contains additional acoustic information about the presence of vowels and other perceptible diacritic indicators. In addition, since diacritics can disambiguate homographs, the presence of diacritics in ASR hy-potheses could potentially lead to different transcriptions. It is possible that the sparsity effects introduced by diacritics would degrade the overall ASR performance, but to what extent is the degradation caused by incorrect diacritics as opposed to incorrect alphabetic characters?

---

[1]The sentence count is obtained from the cleansed edition of the corpus described in [2], which is now the standard corpus for training and evaluating text-based diacritic restoration models.

[2]https://corpus.quran.com/

[3]http://en.arabicspeechcorpus.com

[4]OpenAI's pre-trained ASR model: https://github.com/openai/whisper

Previous research mostly indicate that the presence of diacritics in ASR training hurts ASR performance. However, if we take for granted that diacritized text transcriptions are required for subsequent applications, an increase in overall ASR error rates tells us nothing about the diacritic recognition performance of the model compared with text-based diacritic restoration. Our methodology differs from existing literature in the following aspects: while previous studies evaluated the effect of diacritics on ASR performance, we focus more on evaluating the diacritization performance of the ASR models compared with text-based diacritization as a post-processing step. In addition to reporting overall ASR performance, we isolate the effects of ASR word and character error rate and separately measure diacritics recognition performance using coverage and precision metrics. In our experiments, ASR diacritization significantly outperformed text-based diacritization when the ASR models were trained with manually diacritized transcripts. Using automatic diacritization instead produced mixed results; we observed some performance gains compared to post-processing in some cases, and equivalent results in others.

## 2. Related Work

Al Hani et al. [6] studied the influence of diacritics on the performance of a conventional ASR system using a Pronunciation Mixture Model (PMM) framework [7], a triphone GMM acoustic model, and a trigram language model. The models were trained on 70 hours of speech, and the transcripts were automatically diacritized using a morphological analyzer. In these experiments, modeling diacritics in the lexicon improved performance by 1.7% absolute WER compared to a non-diacritized baseline.

More recent studies generally show the opposite effect, where the inclusion of diacritics in ASR leads to an increase in WER. Abed et al. [8] evaluated eight ASR models (including varieties of GMM and DNN models) with different amounts of training data, with and without diacritics. The largest models were trained on 23 hours of speech. Generally, the inclusion of diacritics reduced the accuracy of the models, but the gap between diacritized and non-diacritized performance gets smaller with more training data. Nevertheless, the authors argue for the benefit of including diacritics in ASR models when integrated with other downstream applications, but they provide no experimental basis for this recommendation.

Alsayadi et al. [9] trained a diacritized end-to-end speech recognition system using 7 hours of transcribed single-speaker data. They reported an overall low WER compared to conventional ASR systems, but did not directly compare diacritized vs. non-diacritized versions. In [10], they trained a non-diacrizied end-to-end ASR model and reported much better performance than the diacritized counterpart. However, they did not evaluate the performance of the diacritization itself and only reported the overall WER of the ASR systems.

## 3. Methodology

In our experiments, we use two recent pre-traiend models that are increasingly adopted in speech applications: Wav2Vec XLS-R [11], and Whisper [12]. We fine-tune each model using a 10-hour single speaker corpus of classical Arabic that has been manually annotated and diacritized, and evaluate the models on a 1-hour held-out test set from the same corpus. We evaluate the following variants of each model:

1. **UD**: UnDiacritized transcripts.
2. **MD**: Manually Diacritized transcripts.
3. **AD**: Automatically Diacritized transcripts.

The UD model is post-processed using a text-based diacritizer to get the final diacritized ASR output. The AD model is pre-processed by removing the gold diacritics and applying text-based automatic diacritization. We experiment with two text-based diacritizers: Shakkelha[5] [13] and the hierarchical deep diacritization[6] D2 model as described in [14] to observe the effect of diacritization error rates on the overall performance.

In order to isolate the overall ASR performance from diacritization performance in particular, we report the following measures:

1. **Unidacritized WER/CER**: ASR word and character error rates, ignoring all diacritics.
2. **Diacritized WER/CER**: Overall ASR error rates including diacritics (the UD model is evaluated with post-added diacritics).
3. Diacritics **Coverage**: the total number of diacritical marks divided by the total number of alphabetic characters.
4. Diacritics **Precision**: the accuracy of diacritization of matching words in ASR hypotheses and references, ignoring no diacritics in the output or the reference[7]. Following conventional practice, we report the precision with and without case ending diacritics, which are the final diacritics for each word. These often correspond to grammatical case, whereas other diacritics correspond to words' morpholigcal structure.

Coverage and precision both measure the diacritization performance of the models, regardless of overall ASR error rates. Since the overall performance is likely to also be affected by the inclusion of diacritics, we report the overall performance in terms of word and character error rates, with and without diacritics.

## 4. Experimental Settings

### 4.1. Data

For training and evaluation, we use the ClArTTS[8] corpus, which is a single-speaker corpus of classical Arabic, manually transcribed with full diacritics. The corpus has about 10 hours of speech for training (9500 short segments), and 1 hour for testing (205 short segments). We also use the Arabic Speech Corpus [3] test set (100 utterances) as an additional out-of-domain set, but due to high ASR error rates caused in part by the unconventional spelling in this set, we focus only on diacritization performance.

---
[5] https://github.com/AliOsm/shakkelha
[6] https://github.com/BKHMSI/deep-diacritization
[7] The standard Diacritic Error Rate (DER) metric used to evaluate text-based diacritizers ignores no-diacritics in references only, and counts no-diacritics in the prediction as errors. Since the latter is included in the coverage metric, we discard both of these cases in our precision metric, and only count the errors where diacritics are present in both reference and prediction.
[8] www.clartts.com

### 4.2. Pre-trained Models & Fine-Tuning

We use the medium pre-trained Whisper[9] model, which is a large pre-trained model for ASR and speech translation, trained on 680K hours of labeled speech data in multiple languages, including Arabic. Without fine-tuning, the model produces mostly undiacritized output. Table 1 shows the performance of the model on the two test set: Classical Arabic TTS Corpus (ClArTTS), and the Arabic Speech Corpus (ASC) [3]. Note that the WER/CER on ASC are rather high due to the unconventional spelling in the corpus as it is annotated for the purpose of speech synthesis. We include this set as an out-of-domain set for the diacritic recognition evaluation. For fine-tuning, we use the original Whisper tokenizer, and fine-tune all model parameters on our training set for 30 epochs.

|           | Corpus  |       |
|-----------|---------|-------|
|           | ClArTTS | ASC   |
| WER       | 16.7%   | 53.4% |
| CER       | 4.8%    | 15.2% |
| Coverage  | 1.9%    | 1.0%  |
| Precision | 18.5%   | 50.0% |

Table 1: *Performance of pre-trained Whisper-medium on the Classical Arabic TTS corpus (ClArTTS) and Arabic Speech Corpus (ASC). Precision is reported w. case ending*

We also use the pre-trained Wav2Vec XLS-R model[10], which is a multilingual model trained on 436K hours of unlabeled speech in 128 languages, including Arabic. The model consists of a CNN feature extractor, followed by a transformer encoder network which is originally trained in a self-supervised manner using contrastive loss. For ASR, we freeze the CNN feature extractor parameters and add a linear layer for classification. The output vocabulary includes all Arabic alphabets and diacritics. We fine-tune the model on our training data for 30 epochs using the CTC loss function [15].

### 4.3. Text-Based Diacritization

The automatically diacritized (AD) models are pre-processed using two text-based diacritizer: D2 and Shakkelha. Table 2 shows the diacritization performance of these models on ClArTTS test set gold transcripts. Note that compared to the reported performance on the Tashkeela corpus, the diacritic error rates are rather high. To make sure this is not merely a result of domain mismatch, we re-trained the D2 model using our training set transcriptions, but the results were worse, possibly due to the training set size, which is orders of magnitude smaller than the Tashkeela corpus. We carried out the remaining experiments using the original pre-trained diacritizers.

|               |          | DER     |          |
|---------------|----------|---------|----------|
| Model         | Coverage | w. case | w.o. case |
| D2 [14]       | 89.8%    | 7.6%    | 6.08%    |
| Shakkelha [13]| 91.5%    | 7.3%    | 6.05%    |
| D2 - retrained| 90.8%    | 9.0%    | 7.19%    |

Table 2: *Performance of text-based diacritization models on ClArTTS test set transcripts*

---

[9] https://huggingface.co/openai/whisper-medium
[10] https://huggingface.co/facebook/wav2vec2-large-xlsr-53

## 5. Results

Table 3 shows ASR overall and diacritization performance for each fine-tuned model on the ClArTTS test set. We report the WER/CER with and without diacritics, in addition to diacritic coverage and precision, with and without case ending diacritics. Note that the coverage of diacritics in the reference transcriptions is 84.6%.

We notice a small change in ASR error rates (excluding diacritics) between the models trained with undiacritized and diacritized transcripts. Including diacritics does change the hypotheses produced by ASR, but the differences in performance are rather small (less than 1% absolute error rate in most cases). Furthermore, the difference is not always in favor of undiacritized ASR; for example, the MD Wav2Vec variants has the lowest character error rate. When it comes to diacritic recognition performance, we see a more significant variations among models. The ASR model fine-tuned with manually diacritized transcripts (MD) achieves remarkably better diacritization performance compared to the models trained with automatically diacritized transcripts (AD). Furthermore, applying text-based diacritization on the output of the undiacritized models (UD +) results in equivalent or worse performance compared to training with automatically diacritized transcripts.

The following listing shows some illustrative examples of the differences in output quality between the diacritized and undiacritized models. We show the gold reference, the output of the Whisper model fine-tuned with manually diacriticed transcripts (MD), and the one fine-tuned without diacritics but post-processed with a text-based diacritizer (UD + D2)[11].

|   |           |                                                     |
|---|-----------|-----------------------------------------------------|
| 1 | Reference | عَلَّمَ وَأَنْشَدَ الرَشِيدُ عَنْ الْمَهْدِي         |
|   | MD        | عَلَّمُوا وَأَنْشَدَ الرَشِيدِ عَنْ الْمَهْدِي       |
|   | UD + D2   | عِلْمٌ وَأَنْشَدَ الرَّشِيدُ عَنْ الْمَهْدِي         |
| 2 | Reference | انْفَرِدْ بِسِرِّكَ وَلَّا تُودِعْهُ حَازِمًا        |
|   | MD        | امْفَرِدْ بِسِرِكَ وَلَا تُذِعْهُ حَازِمًا           |
|   | UD + D2   | انْفَأَرْتَ بِسِرِكِ وَلَا تَذَعُهُ حَازِمًا         |
| 3 | Reference | يُفْسِدُ مَا حَوْلَهُ لَكِنْ اتَّبَعْتُ فِيهِمْ      |
|   | MD        | يُفْسِدُ مَا حَوْلَهُ لَكِنْ اتَّبَعْتُ فِيهِمْ      |
|   | UD + D2   | يُفْسِدُ مَا حَوْلَهُ لَكِنْ أَتَّبَعَتْ فِيهِمْ     |

Example 1 shows a case where the MD model produces an incorrect first word while the UD model produces the correct transcript. Note that the word in this context is ambiguous. However, the diacritization of the MD model corresponds to the way the word sounds in the reference text. The UD output, after being processed with D2, results in incorrect diacritics. Example 2 is a case where both MD and UD outputs have errors. We see that in spite of the incorrect characters, the MD model produces diacritics the reflect the way the original words sound, whereas the D2 diacritizer produces diacritics that seem sensible without additional context, but do not actually reflect the original words. Example 3 shows a case where both MD and UD produce the correct characters, but the D2 text diacritizer results in a different conjugation of the second verb. Without additional textual context, there is no way to identify the correct diacritics in this case, but the MD model produces the correct output as it corresponds to the audio signal.

---

[11] In these examples, gemination diacritics are not shown due to the specific encoding scheme used in LATEX

|  |  | Without Diacritics | | With Diacritics | | | | |
|  |  | WER | CER | WER | CER | Coverage | Precision | |
|  |  |  |  |  |  |  | w. case | w.o. case |
| --- | --- | --- | --- | --- | --- | --- | --- | --- |
| Wav2Vec | UD + D2 | **8.1%** | 2.2% | 38.5% | 9.5% | 84.3% | 94.08% | 95.37% |
|  | UD + Shakkelha | **8.1%** | 2.2% | 39.3% | 9.6% | 84.6% | 93.28% | 94.74% |
|  | MD | 8.4% | **2.1%** | **16.0%** | **3.0%** | 84.5% | **98.26%** | **99.16%** |
|  | AD : D2 | 9.6% | 5.5% | 42.0% | 10.6% | 83.8% | 93.98% | 95.83% |
|  | AD: Shakkelha | 8.9% | 2.3% | 40.6% | 9.2% | 84.5% | 93.99% | 95.47% |
| Whisper | UD + D2 | **6.4%** | **1.8%** | 38.4% | 9.3% | 84.6% | 93.88% | 95.17% |
|  | UD + Shakkelha | **6.4%** | **1.8%** | 40.1% | 10.0% | 84.6% | 93.16% | 94.53% |
|  | MD | 6.5% | 1.9% | **13.4%** | **2.8%** | 84.6% | **98.42%** | **99.08%** |
|  | AD : D2 | 6.7% | 2.1% | 38.4% | 8.9% | 83.9% | 94.23% | 95.44% |
|  | AD: Shakkelha | **6.4%** | **1.8%** | 36.4% | 8.7% | 84.7% | 95.04% | 96.38% |

Table 3: *Performance of fine-tuned ASR models on ClArTTS test set in terms of Word Error Rate (WER), Character Error Rate (CER), diacritic Coverage and Precision. MD: manually diacritized training data. UD: undiacritized. AD: automatically diacritized. We show the diacritization models used to post-process UD or pre-process AD.*

Since the results reported in Table 3 are based on test data drawn from the same audiobook as the training set, and both are annotated following the same guidelines, we use the Arabic Speech Corpus as an out-of-domain test set to verify whether the same patterns of performance hold in different speech and annotation conditions. The results for the Whisper model are shown in table 4. We do not report ASR error rates since they are similar to the pre-trained model, and these error rates generally don't reflect true ASR performance due to spelling mismatches. Since the precision metric relies only on matching words in ASR hypotheses and the reference transcriptions, the results shown in table 4 reflect the diacritic recognition performance of the models regardless of ASR performance. Consistent with the results on in-domain test set, we see that manually-diacritized training data lead to higher precision compared with automatically diacritized data. Furthermore, we see a larger gap between ASR diacritic performance and text-based diacritic performance. In this domain, the text-based diacritizers achieve much lower precision compared with ASR diacritization, even when compared to models trained with automatically diacritized data.

|  | Coverage | Precision | |
|  |  | w. case | w.o. case |
| --- | --- | --- | --- |
| UD + D2 | 82.3% | 85.47% | 89.97% |
| UD + Shakkelha | 82.3% | 84.68% | 87.83% |
| MD | 83.3% | 96.55% | 98.32% |
| AD : D2 | 82.3% | 92.94% | 95.93% |
| AD : Shakkelha | 83.0% | 91.03% | 93.87% |

Table 4: *Diacritic recognition performance of fine-tuned Whisper models on the Arabic Speech Corpus Test Set*

## 6. Discussion

We carried out experiments to target the diacritic recognition performance of Arabic ASR systems regardless of ASR error rates. We fine-tuned pre-trained models using a manually transcribed corpus with diacritics to encourage ASR models to produce diacritics for most characters. Our baseline for comparison is a fine-tuned model trained with non-diacritized transcriptions and post-processed using text-based diacritic restoration models.

All models resulted in high coverage of diacritics close to the reference coverage rate. In terms of precision, models fine-tuned with manually diacritized data resulted in higher precision compared to all other variants. In our in-domain test set, we observed little difference in performance between ASR models trained with automatic diacritization compared with text-based diacritization as a post-processing step. However, in out-of-domain data, we observed a larger difference in favor of ASR diacritization, regardless of whether the data is manually or automatically diacritized. One possible explanation is that the predicted output in the out-of-domain data set is rather erroneous, resulting in many misspelled words. This could potentially confuse the text-based diacritizers, which rely more on word identity and surrounding context words. In general, we noticed that the diacritized ASR models produce diacritics more consistent with the sound of the words, even in the presence of unknown words and errors. The undiacritized ASR models, on the other hand, produce ambiguous output that can be difficult to disambiguate directly from text. In addition, speech is generally less structured than text, so a diacritization model trained on text would not necessarily generalize to the speech domain. We observed a significant reduction in diacritization accuracy even when using the gold transcriptions as input to the pre-trained text-based diacritizers. Further evaluations in other test domains are needed to support these conclusions, but our results in this paper indicate that ASR diacritization have higher potential in terms of diacritic recognition accuracy, and could generalize to domains unseen in training.

One of the limitations in this work is the relatively small size of the speech corpora used for training and evaluation, which is due to the shortage of public corpora with fully diacritized transcripts. The datasets used for this work were curated for the purpose of text-to-speech synthesis, so they have been manually diacritized, but they contain single-speaker recordings with little variation in tone and environmental factors. This limited our capacity to test our conclusions on more general and varied conditions. However, since we start with pre-trained models, the fine-tuned systems can generally handle new speech conditions rather well, at least when compared with the pre-trained versions. Furthermore, as we mainly focused on evaluating ASR diacritization performance against text-based diacritizers, ideally we would have used the same text data to train both models to ensure consistency of annotations and domain. We attempted to re-train the text-based diacritizers using the transcriptions of the training set to have a fair comparison, but the results were worse than the diacritizers trained on Tashkeela, which is a much larger text corpus. On the other hand, the same point may indicate a higher potential for ASR diacritization as it can generalize from a smaller set of examples. It is likely that the ASR models learned to map acoustic features to diacritics in addition to other contextual cues, which leads to higher accuracy for unseen words or novel contexts. The results on the out-of-domain dataset also support the same conclusions as we achieved the best performance using the model trained with manually diacritized data.

In terms of the underlying ASR performance, adding diacritics in the training set transcriptions does not seem to hurt performance; we did observe a generally higher WER/CER but the difference is rather small and could potentially be rectified with a slightly larger corpus. Since large, manually diacritized speech corpora are generally unavailable, fine-tuning on a smaller corpus as done in this work is a viable approach that leads to robust diacritization performance.